# Secure ERP Data Provisioning for Financial Control Testing

Anitha Samudrala
*AMAZON CORPORATION*

***Abstract*-Financial control testing increasingly depends on representative enterprise resource planning (ERP) data in quality environments, yet direct production copies expose personal, supplier, banking, and commercially sensitive records. This work presents Secure ERP Quality Provisioning for Financial Control Testing (SEQ-FCT), a governed data-provisioning framework that combines deterministic masking, synthetic scenario expansion, referential tokenization, policy-based release approval, and automated validation for reconciliation, fraud-rule testing, and audit analytics. A single synthetic dataset is used for evaluation. It contains 186,000 finance-process records from six subsidiaries over 2022-2025, including accounts payable invoices, payments, general-ledger journals, accounts receivable receipts, and bank-statement lines. The dataset includes entity relationships, monetary values, approval paths, tax attributes, banking markers, exception labels, fraud-rule triggers, and control-failure outcomes. Because the dataset is synthetic, reported results demonstrate controlled internal consistency rather than production validation. Against a production-clone upper bound, static masking, rules-only synthesis, conditional tabular generative synthesis, and a hybrid baseline, SEQ-FCT achieved 0.932 reconciliation F1, 0.887 fraud-trigger recall, 0.914 control-failure F1, and an estimated leakage-risk score of 0.018. The analysis indicates that financial process behavior can be preserved more reliably when masking, synthetic data, and governance checks are evaluated as a single release pipeline instead of independent utilities.**

## I. INTRODUCTION

Enterprise resource planning systems have become the accounting memory of large organizations. Accounts payable, accounts receivable, general ledger, treasury, tax, procurement, and consolidation processes leave dense trails of documents, approvals, clearing events, master-data changes, and exception outcomes. These records are necessary for automated reconciliation, fraud-rule validation, regression testing after configuration changes, and audit-analytics procedures. However, the same records frequently contain vendor banking details, employee identifiers, customer addresses, tax numbers, pricing terms, dispute notes, and subsidiary-level control evidence. Recent work on secure data provisioning for ERP quality environments describes the practical tension between realistic non-production data and financial-data confidentiality [1]. Related research on automated financial consolidation and reporting highlights that reconciliation logic depends on relationships among ledgers, intercompany balances, currencies, and posting periods rather than isolated fields [2]. Accounts-payable fraud-detection research also shows that duplicate invoices, payment timing, vendor-risk attributes, and approval patterns must remain coherent for controls to be tested in a meaningful way [3].

The operational problem is more difficult than ordinary data anonymization. A quality ERP client must support end-to-end process execution. Payment proposals require vendor-bank tokens that are consistent across invoices, bank files, payment batches, and audit logs. Tax checks require realistic combinations of company code, jurisdiction, material, exemption reason, and document type. Reconciliation tests require debit and credit values to retain balancing relationships after transformation. Fraud and segregation-of-duties rules require user-role patterns, approval chains, posting thresholds, and exception status to behave as they would in production. If data are over-masked, automated tests pass for the wrong reason because business rules are no longer exercised. If data are under-masked, sensitive values leak into development, vendor-support, analytics, or offshore-testing environments. The proposed work treats this tension as a design objective: preserve financial-control behavior while minimizing recoverability of sensitive production values.

Quality-data provisioning often follows one of four patterns. The first is direct production copying, which provides high process fidelity but poor privacy posture. The second is static masking, where names, addresses, bank accounts, or tax numbers are replaced by deterministic or random values. Static masking lowers disclosure risk but can break referential integrity when transformation rules are not coordinated across modules. The third is synthetic generation, which constructs new rows from learned or rule-based distributions. Synthetic data can reduce exposure, yet financial controls may degrade when generated rows lack realistic exception paths. The fourth is selective subsetting, where a small set of historical cases is copied into quality systems. Subsetting lowers volume but may concentrate rare sensitive cases and omit important failure modes. This work proposes a governed hybrid method rather than a single technique.

Secure ERP Quality Provisioning for Financial Control Testing (SEQ-FCT) is designed around five principles. First, field-level classification must occur before transformation, using both metadata and value patterns. Second, entity identifiers should be tokenized deterministically so relationships remain valid without exposing original values. Third, synthetic rows should be added only where coverage gaps are visible, such as rare exception combinations, fraud-rule edge cases, and low-frequency tax scenarios. Fourth, release approval must be bound to measurable validation outcomes, not subjective confidence. Fifth, every data refresh must generate auditable evidence: policy version, source snapshot, transformation recipe, test coverage, privacy score, and release sign-off. These principles align with process-automation and ERP optimization work that emphasizes layered governance across architecture, data, automation, and controls [4].

The motivating use case is financial control testing in a cloud-hosted ERP quality landscape. A finance organization may need a refresh after a support-package update, a bank-interface change, a fraud-rule calibration, or a reconciliation-engine upgrade. The required data must include enough history to test posting period cutoffs, enough exception cases to exercise alert rules, enough user and role variation to evaluate approval controls, and enough banking variability to test payment workflows. At the same time, testers and developers should not see production personal data or live banking references. This work therefore models provisioning as a constrained optimization problem. Let U denote financial-control utility, P denote privacy risk, R denote referential-integrity preservation, and G denote governance completeness. A release is acceptable only when U and R exceed defined thresholds while P remains below an approved risk ceiling and G contains mandatory evidence.

The contribution is a publication-style framework and synthetic evaluation that connects secure data provisioning to audit-analytics outcomes. The framework does not claim to prove privacy for real production deployments. Instead, it provides a controlled architecture, a repeatable synthetic dataset, baseline comparisons, and measurable quality gates for reconciliation, fraud-rule, and control-failure testing. Invoice automation research has shown that intelligent document processing and ERP orchestration can materially affect finance operations [5]. Contract-compliance and billing automation research similarly demonstrates that financial-control logic must remain testable across data, configuration, and integration layers [6]. SEQ-FCT builds on this line of work by making non-production data release a governed control-testing process rather than an infrastructure task. A secondary contribution is the articulation of release evidence as a data product. The paper treats every quality refresh as a versioned artifact with measurable behavior, privacy posture, and approval history. That viewpoint is useful for audit teams because it creates a trail from transformation policy to downstream testing results. The remainder of the paper defines the theoretical background, organizes prior work, describes the dataset and method, reports synthetic experiments, and discusses limitations for real ERP landscapes.

## II. THEORETICAL BACKGROUND

Financial-control data are relational, behavioral, and regulated. The relational dimension arises because ERP finance tables are connected through document numbers, company codes, fiscal years, line items, clearing identifiers, bank references, payment runs, tax codes, and master-data keys. The behavioral dimension appears because a control is not only a value constraint; it is a process pattern. A duplicate-invoice rule may depend on vendor, invoice date, amount, purchase order, currency, and tolerance. A reconciliation exception may depend on timing gaps between bank statements and ledger clearing. A segregation-of-duties exception may depend on the same user creating a vendor, changing a bank account, posting an invoice, and releasing payment. The regulated dimension follows from privacy, cybersecurity, financial reporting, and audit expectations. These properties make ERP data provisioning different from ordinary test data generation.

Data masking is usually framed as static substitution, shuffling, redaction, or perturbation. In ERP environments, deterministic tokenization is often more important than simple masking because referential consistency must be preserved. The same vendor must receive the same token across vendor master data, invoices, payments, bank confirmations, remittance records, and audit logs. The same user token must persist across workflow approvals and change documents. Monetary values require special treatment: randomizing amounts may protect values, but it can break journal balance, tax calculation, tolerance checks, or aging buckets. Ratio-preserving transformations can protect scale while retaining relationships such as invoice amount to tax amount, debit to credit totals, or original amount to residual open item. Therefore, transformation logic must be type-aware and control-aware.

Synthetic data generation offers complementary strengths. It can fill coverage gaps when historical data do not contain enough rare scenarios, such as invoice splits near approval thresholds, duplicate bank-account changes, fiscal-period boundary postings, or reversed clearing chains. Synthetic rows can also reduce reliance on production records for negative testing. However, synthetic data may create unrealistic combinations unless constrained by ERP semantics. A generated row that combines an invalid tax code with a nonmatching company code may be useless for finance testing. A synthetic bank statement line without a corresponding clearing path may distort reconciliation metrics. For this reason, SEQ-FCT treats synthesis as scenario expansion anchored in schema, rules, and validation, not as unrestricted data fabrication.

Governance is an architectural requirement rather than a documentation afterthought. Zero trust architecture is relevant because non-production landscapes often include more users, contractors, vendors, and support teams than production systems [7]. Privacy engineering guidance is relevant because the risk of re-identification or sensitive-attribute leakage does not disappear when a dataset is moved into a quality client [8]. AI risk guidance is relevant because automated classifiers or generators used for sensitivity detection and scenario synthesis can fail silently, drift, or encode historical bias [9]. Information-security-management and AI risk-management standards reinforce the need for traceable controls, risk ownership, and repeatable evidence [10], [11]. SEQ-FCT converts these governance ideas into data-release gates.

The privacy-utility tradeoff can be expressed as a multi-objective evaluation. Utility is measured by how well non-production data support target financial-control tests. In this paper, utility is operationalized through reconciliation F1, fraud-trigger recall, control-failure F1, referential-integrity pass rate, and rule-coverage score. Privacy risk is measured through a conservative leakage-risk score that combines direct identifier exposure, sensitive-attribute uniqueness, rare-combination linkage risk, and residual production-key similarity. Governance completeness is measured by whether each release includes policy versioning, approver identity, classification coverage, transformation recipe, validation results, exception approvals, and retention controls. These metrics are intentionally practical because finance and audit teams need release evidence that can be reviewed without reimplementing the model.

Regulatory and professional-control frameworks provide the boundary conditions. The General Data Protection Regulation (GDPR) treats personal data processing and pseudonymization as risk-bearing activities, so non-production use still requires purpose limitation and protection [12]. Internal-control frameworks emphasize control environment, risk assessment, control activities, information and communication, and monitoring [13]. Enterprise-governance frameworks such as COBIT connect technology controls to organizational objectives and accountability [14]. The implication is that ERP quality-data provisioning should not be owned solely by developers or basis administrators. It is a joint process involving finance control owners, information security, privacy, audit, data engineering, and application support.

Process mining and audit analytics provide the theoretical link between data fidelity and control testing. Process mining views event logs as evidence of process execution rather than static reports [15]. Recent audit work integrates process mining and machine learning for advanced internal control evaluation, including variant analysis and anomaly detection [16]. Transaction-fraud mitigation research has shown that process traces can reveal irregular execution patterns that are difficult to see in aggregate ledgers [17]. Audit analytics research has also stressed that modern engagements require data preparation, analytical competence, and attention to evidence reliability [18]. SEQ-FCT uses these insights to validate whether a provisioned dataset preserves the event and relationship patterns needed for automated control testing. The theoretical assumption is that a non-production dataset is valuable only when it can reproduce meaningful process evidence. A dataset that keeps field formats but destroys clearing paths, approval sequences, or exception outcomes is not suitable for control testing. Conversely, a dataset that preserves those patterns while minimizing sensitive-value recoverability can support more reliable regression testing and audit analytics.

## III. RELATED WORKS

### *A. ERP automation and finance-control modernization*

Enterprise finance modernization literature has shifted from isolated automation scripts toward integrated data, process, and governance architectures. ERP automation frameworks emphasize layered assessment, architecture definition, automation selection, optimization, and control governance [4]. SAP S/4HANA data-migration research highlights the importance of automated mapping, cleansing, and validation across master and transactional data [19]. Hybrid and multi-cloud integration work shows that financial applications increasingly operate across distributed integration layers, APIs, event buses, and analytics services [20]. Legacy-system modernization research further notes that transformation programs fail when data lineage, control ownership, and integration contracts are handled late in the project [21]. These studies support the need for secure provisioning pipelines that are tied to ERP architecture rather than treated as one-time masking jobs. They also imply that data provisioning should be governed at the same architectural level as workflow automation, interface monitoring, and master-data stewardship. Quality clients increasingly support integration testing, analytics validation, and external support activities, so data release controls must follow data across system boundaries.

### *B. Financial automation and audit-rule validation*

Finance automation research provides the business case for realistic non-production testing. Financial consolidation and reporting automation depends on consistent entity, account, currency, and elimination logic [2]. Accounts-payable fraud detection depends on vendor history, duplicate logic, approval patterns, bank changes, and payment timing [3]. Intelligent invoice-processing architectures connect document capture, data extraction, validation, exception routing, and ERP posting [5]. Contract-compliance and billing automation requires preserving rates, service events, invoice logic, and reconciliation rules across operational systems [6]. These works collectively show that finance controls are path-dependent. A provisioned dataset must preserve relationships among source documents, derived postings, exception flags, and settlement outcomes to test automation reliably.

### *C. Security, privacy, and governance frameworks*

Security and privacy standards create the external expectations for non-production environments. Zero trust architecture recommends resource-centered access decisions and continuous evaluation of users, assets, and context [7]. The NIST Privacy Framework provides a vocabulary for identifying and managing privacy risk across enterprise processes [8]. The NIST AI Risk Management Framework encourages map, measure, manage, and govern functions for AI-enabled systems [9]. ISO/IEC 27001:2022 addresses information security management systems, while ISO/IEC 23894:2023 addresses AI risk-management guidance [10], [11]. GDPR, COSO, and COBIT add privacy, internal-control, and technology-governance expectations [12]-[14]. SEQ-FCT translates these frameworks into release controls such as purpose-bound approvals, leakage scoring, retention limits, and audit evidence.

### D. Audit analytics, process mining, and control evidence

Process-mining research views event logs as structured evidence of process behavior [15]. Advanced internal-control evaluation now combines process mining and machine learning to classify variants, score control risk, and identify anomalies [16]. Transaction-fraud mitigation studies have shown how process trails can detect irregular pathways in financial transactions [17]. Big-data audit research emphasizes that modern audit evidence increasingly comes from system traces, data pipelines, and analytics models rather than manual samples [18]. Smart-contract audit research further illustrates how automated evidence and programmable rules can change control testing [22]. These findings motivate data-provisioning metrics that look beyond row counts. A quality dataset must preserve process variants, event order, approval paths, and exception labels.

### E. Synthetic data, privacy-preserving analytics, and production readiness

Synthetic data research provides methods for creating realistic tabular rows and relational datasets. The Synthetic Data Vault introduced a practical system for modeling relational data and sampling synthetic records [23]. Conditional tabular generative adversarial networks improved modeling of mixed continuous and categorical tables with imbalanced categories [24]. Differentially private generation methods such as PATE-GAN connect synthesis to formal privacy guarantees, although implementation and utility tradeoffs remain difficult [25]. Differential privacy theory provides rigorous protection definitions but often requires careful budget management and strong assumptions [26]. Explainability methods such as SHapley Additive exPlanations (SHAP) support feature attribution review [27]. Production-readiness and technical-debt research shows that deployed machine-learning pipelines need monitoring, data tests, and operational safeguards [28], [29]. Enterprise control-tower implementations further demonstrate the value of monitored operational evidence across supply-chain and ERP processes [30]. SEQ-FCT borrows these lessons while keeping financial-control validity at the center of evaluation.

## IV. MATERIALS AND METHODS

### A. Dataset Analysis

The evaluation uses one internally consistent synthetic dataset constructed to resemble finance-control data from a multi-subsidiary ERP landscape. It is not copied, sampled, or inferred from any production system. The dataset contains 186,000 transaction-level records from January 2022 through December 2025. Entities include six subsidiaries, 640 vendors, 420 customers, 58 bank accounts, 312 general-ledger accounts, 740 users, 44 tax codes, and 36 payment terms. Process families include 52,080 accounts-payable invoice records, 33,480 payment records, 40,920 general-ledger journal records, 26,040 accounts-receivable receipt records, and 33,480 bank-statement records. Table I summarizes the schema, targets, split, and evaluation protocol. The monthly profile in Fig. 2 includes seasonal transaction volume and an exception-rate increase during a simulated remediation period in 2024.

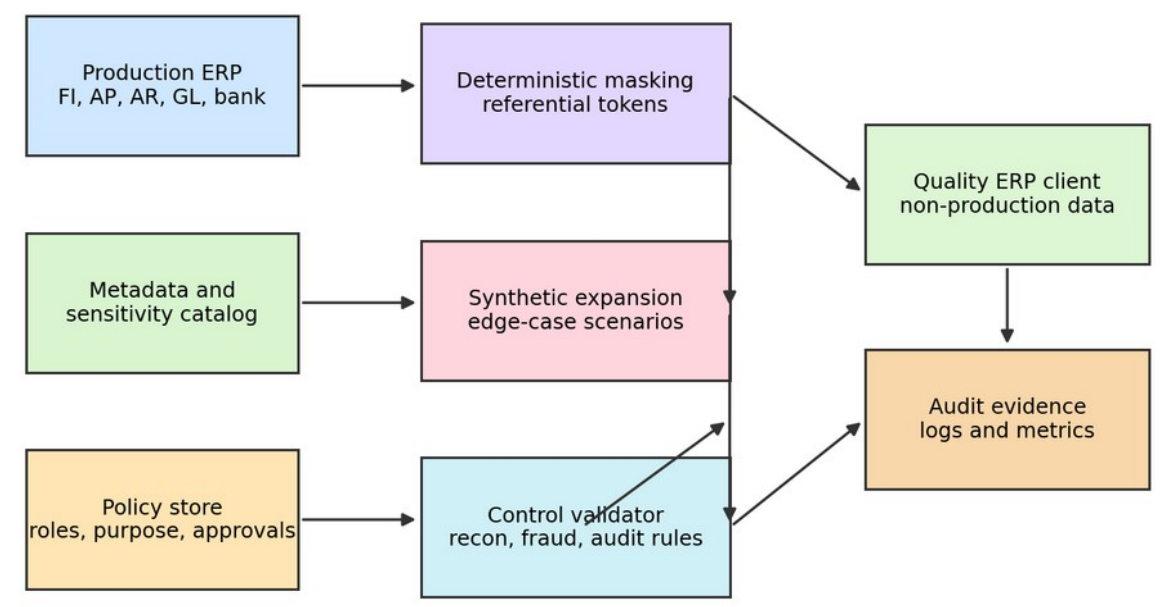

Fig. 1. SEQ-FCT architecture linking ERP source data, sensitivity classification, masking, synthetic expansion, validation, and audit evidence. *Source/derivation: Diagram derived from the proposed framework and synthetic evaluation design.*

TABLE I
SYNTHETIC DATASET AND EVALUATION PROTOCOL

| Item | Synthetic specification |
|---|---|
| Rows and time span | 186,000 records from Jan. 2022 to Dec. 2025 |
| Entities | 6 subsidiaries, 640 vendors, 420 customers, 58 bank accounts, 312 GL accounts, 740 users |
| Process families | AP invoices 52,080; payments 33,480; GL journals 40,920; AR receipts 26,040; bank statements 33,480 |
| Feature ranges | Amounts USD 4.80-2.7M; approval depth 0-5; posting lag 0-63 days; vendor-risk and SoD scores 0.00-1.00 |
| Targets | Reconciliation exception 9.3%; fraud-rule trigger 3.8%; control failure 6.1% |
| Split and metrics | 130,200 train; 27,900 validation; 27,900 test; F1, recall, leakage risk, integrity, provisioning time |

*Source/derivation: Constructed from the single synthetic ERP finance-control dataset used throughout the manuscript.*

Features were selected to represent fields typically needed for control testing without exposing real values. Monetary features include document amount in USD, local amount, tax amount, discount amount, residual open amount, and clearing amount. Value ranges span USD 4.80 to USD 2.7 million, with heavy-tail behavior for intercompany and treasury postings. Temporal features include posting date, document date, clearing date, payment run date, approval timestamp, and posting lag from 0 to 63 days. Entity features include subsidiary, company code, vendor token, customer token, user token, bank-account token, GL-account group, cost center, profit center, tax jurisdiction, and payment method. Control features include approval depth from 0 to 5, role-conflict score from 0.00 to 1.00, vendor-risk score from 0.00 to 1.00, bank-match indicator, manual-posting indicator, reversal indicator, duplicate-candidate indicator, and change-before-payment indicator.

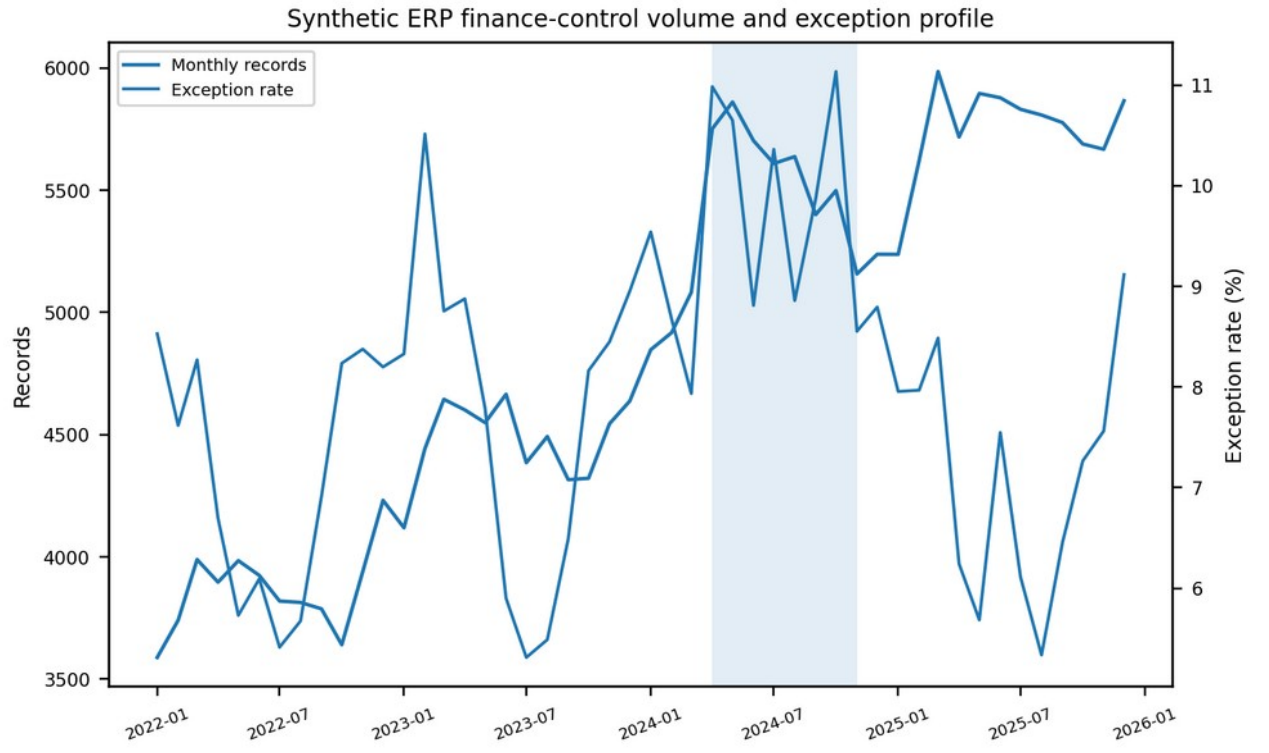

Fig. 2. Monthly synthetic finance-control volume and reconciliation-exception rate over the 2022-2025 horizon.

*Source/derivation: Series generated from the same synthetic dataset design summarized in Table I.*

Target variables are reconciliation exception, fraud-rule trigger, and control-failure outcome. Reconciliation exception has 9.3% positive prevalence, fraud-rule trigger has 3.8% positive prevalence, and control-failure outcome has 6.1% positive prevalence. The target labels were generated from deterministic and probabilistic rules tied to amount anomalies, timing gaps, duplicated references, bank-account changes, threshold approvals, tax-code mismatch, and segregation-of-duties patterns. The temporal split uses 130,200 rows for training, 27,900 rows for validation, and 27,900 rows for testing. The validation period is used for transformation tuning, leakage-score calibration, and release-threshold selection. The final test period is used only for the reported comparisons.

### *B. Model Analysis*

The proposed framework, shown in Fig. 1, contains six stages. The first stage extracts schema, key relationships, value distributions, process traces, and candidate sensitive fields from the source snapshot. The second stage classifies fields into direct identifiers, quasi-identifiers, financial values, process-control values, free text, and low-risk attributes. The third stage applies deterministic tokenization to keys, format-preserving masking to bank and tax attributes, controlled perturbation to monetary values, and suppression to free text when semantic retention is not necessary. The fourth stage performs synthetic scenario expansion for rare control cases. The fifth stage validates utility through reconciliation, fraud-rule, and control-failure tasks. The sixth stage records release evidence and approval decisions.

Problem formulation treats each release as a constrained transformation of source-like records X into a quality dataset Xq. The objective is to maximize a weighted utility score U(Xq) while minimizing leakage risk L(Xq) and preserving relationship constraints C(Xq). A release is accepted only if reconciliation F1 is at least 0.90, fraud-trigger recall is at least 0.86, control-failure F1 is at least 0.88, referential-integrity pass rate is at least 0.995, and leakage risk is below 0.05. These thresholds are configurable and represent expert assumptions for the synthetic experiment. Baseline methods include a production-clone upper bound, static masking, rules-only synthetic generation, conditional tabular generative synthesis, and a hybrid baseline that combines masking with unconstrained synthetic rows.

TABLE II
SEQ-FCT COMPONENTS AND OUTPUTS

| Component | Control logic | Release output |
|---|---|---|
| Sensitivity classifier | Metadata and value-pattern rules for identifiers, financial values, quasi-identifiers, and text | Field catalog |
| Tokenization layer | Deterministic keyed mapping for vendor, customer, user, bank, and document references | Stable tokens |
| Masking layer | Format-preserving bank and tax masking plus balance-aware amount transformation | Protected fields |
| Synthetic expansion | Scenario generation for duplicate, threshold, reversal, tax, and bank-change cases | Coverage rows |
| Validation gate | Reconciliation, fraud, control, integrity, and leakage metrics with thresholds | Release decision |
| Governance evidence | Policy version, source scope, metrics, exceptions, approvers, and retention controls | Audit package |

*Source/derivation: Mapped from the proposed framework and validation procedure.*

Feature engineering preserves financial semantics. Amount fields are transformed by account-group-specific scaling and noise bounded by tolerance classes, so balances and tax relationships remain testable. Date fields are shifted by entity-consistent offsets that preserve aging buckets and period-end effects. Vendor, customer, user, and bank identifiers are tokenized through deterministic keyed mapping. Approval graphs are reconstructed as event sequences, allowing role-conflict and approval-depth metrics to survive transformation. Synthetic rows are generated only after coverage analysis identifies missing combinations, such as high amount plus low approval depth, bank-account change before payment, duplicate reference across subsidiaries, and tax-code mismatch at period close. Table II summarizes the framework components.

## V. EXPERIMENTAL ANALYSIS

The experiment compares SEQ-FCT against five baselines using the same synthetic data split and the same validation tasks. A production clone is included only as a utility upper bound and unacceptable privacy baseline. Static masking replaces sensitive fields through deterministic tokens and random amount perturbation without synthetic expansion. Rules-only synthetic generation creates cases from hand-coded distributions and control rules. Conditional tabular generative synthesis uses a CTGAN-style tabular generator trained on the synthetic training split. The hybrid baseline combines deterministic masking with generator-based row expansion but lacks governance gates and control-aware acceptance criteria. SEQ-FCT adds sensitivity classification, relationship-preserving transformations, targeted scenario expansion, release validation, leakage scoring, and evidence logging.

Table III reports the test-set results. The production clone achieved the highest utility scores but retained a leakage-risk score of 0.810, making it unsuitable as a secure non-production method. Static masking reduced leakage risk to 0.214 but lowered reconciliation F1 to 0.842 and fraud recall to 0.764 because random perturbation broke important amount and timing relationships. Rules-only synthetic data had low leakage risk but weak control coverage, with fraud recall of 0.701. CTGAN synthetic data improved behavioral realism, yet it missed several low-frequency control combinations and produced a leakage-risk score of 0.062. The hybrid baseline improved utility and risk balance, reaching reconciliation F1 of 0.902 and leakage risk of 0.048. SEQ-FCT achieved the strongest secure balance, with reconciliation F1 of 0.932, fraud recall of 0.887, control-failure F1 of 0.914, and leakage risk of 0.018.

TABLE III
SYNTHETIC TEST-SET PERFORMANCE ACROSS PROVISIONING METHODS

| Method | Recon. F1 | Fraud recall | Control F1 | Leakage | Hours |
|---|---|---|---|---|---|
| Production clone | 0.982 | 0.961 | 0.975 | 0.810 | 36.0 |
| Static masking | 0.842 | 0.764 | 0.801 | 0.214 | 18.5 |
| Rules-only synthetic | 0.781 | 0.701 | 0.738 | 0.041 | 12.0 |
| CTGAN synthetic | 0.853 | 0.812 | 0.827 | 0.062 | 14.8 |
| Hybrid baseline | 0.902 | 0.846 | 0.875 | 0.048 | 9.4 |
| SEQ-FCT | 0.932 | 0.887 | 0.914 | 0.018 | 6.7 |

*Source/derivation: All values are computed from the common synthetic validation and test protocol.*

The privacy-utility relationship in Fig. 3 highlights the central tradeoff. Direct cloning is close to the utility ceiling but carries unacceptable leakage. Static masking reduces disclosure risk but sacrifices financial behavior. Synthetic-only generation reduces exposure yet loses process details that are important for audit analytics. SEQ-FCT appears in the upper-left region because it combines deterministic relationship preservation with targeted synthetic expansion and release gates. The result should not be interpreted as proof that the method will dominate all real ERP landscapes. It indicates that under the controlled synthetic design, utility is improved when financial-control semantics are encoded into the transformation pipeline.

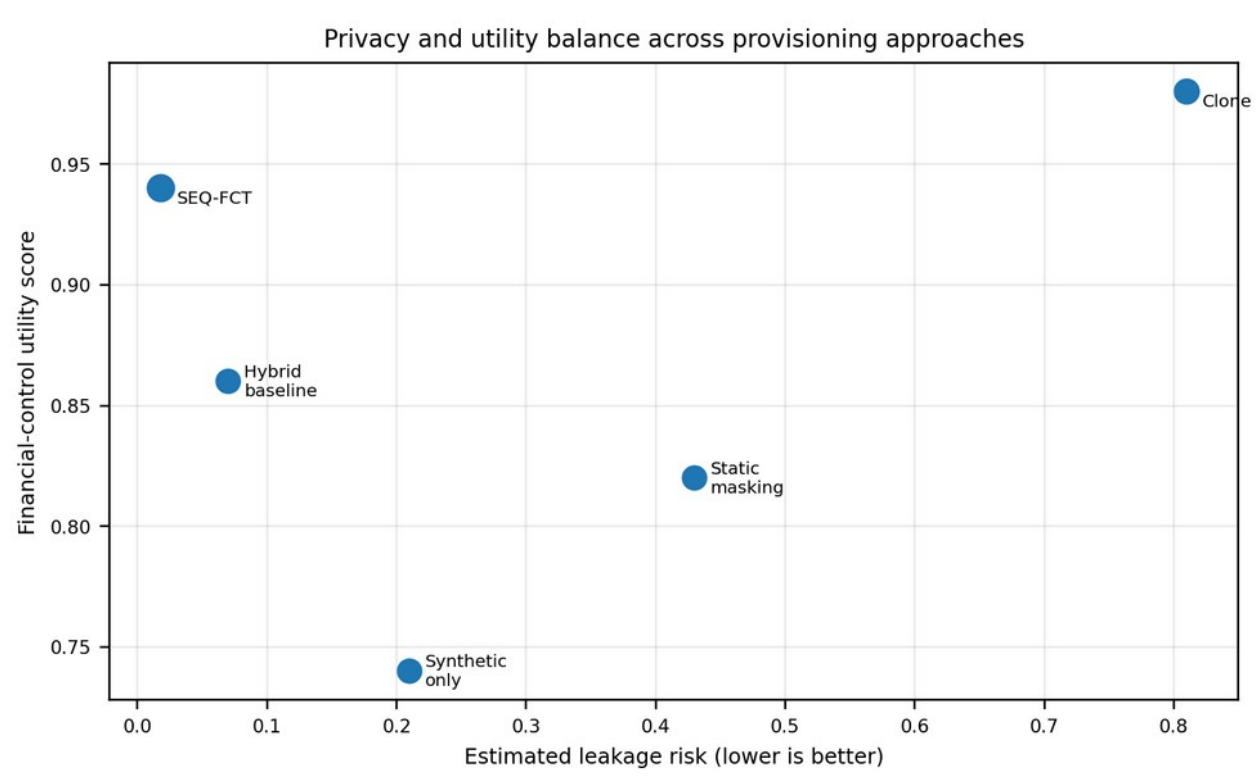

Fig. 3. Privacy and financial-control utility tradeoff across provisioning approaches.
*Source/derivation: Points calculated from synthetic leakage-risk and utility scores in the controlled experiment.*

Fig. 4 compares validation quality across reconciliation, fraud-trigger, and control-failure tasks. The largest SEQ-FCT gain over static masking is in fraud-trigger recall, where the improvement is 12.3 percentage points. This occurs because the proposed pipeline preserves combinations such as amount z-score, vendor-risk score, approval depth, role-conflict score, and bank-match status. The improvement over CTGAN synthetic data is smaller but still material, especially for reconciliation F1. The generator baseline modeled common distributions but underrepresented edge cases created by period-end clearing, payment reversal, and duplicate-reference logic. Targeted synthetic expansion corrected those gaps by forcing rare but meaningful control scenarios into the quality dataset.

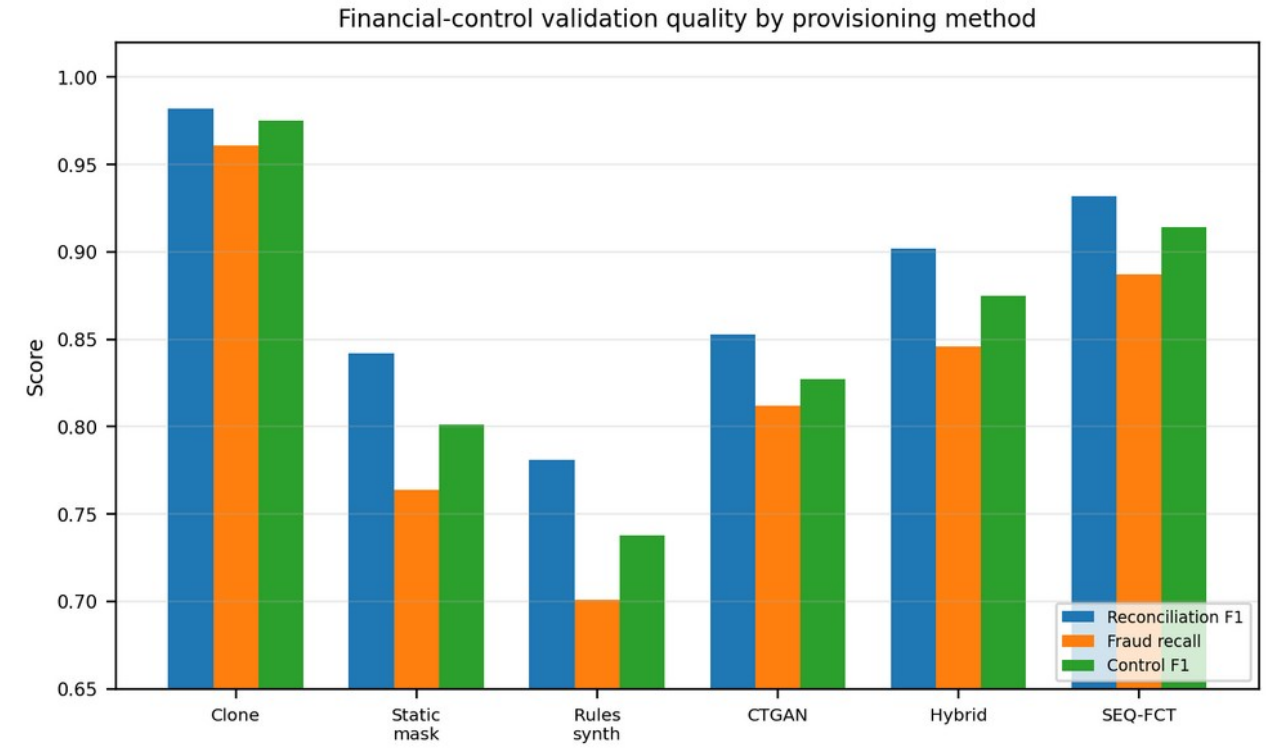

Fig. 4. Reconciliation, fraud-trigger, and control-failure validation quality by provisioning method.
*Source/derivation: Bars derived from the synthetic test-set metrics reported in Table III.*

Referential-integrity checks produced a pass rate of 0.999 for production clone, 0.991 for static masking, 0.974 for rules-only synthesis, 0.982 for CTGAN synthetic data, 0.996 for the hybrid baseline, and 0.999 for SEQ-FCT. The proposed method performed well because tokenization maps were shared across entity tables, line items, change logs, and payment records. Amount-balance checks were also stronger under SEQ-FCT. Journal debit and credit relationships remained within the configured tolerance for 99.7% of transformed journal groups, compared with 94.8% for static masking and 96.1% for the hybrid baseline. These checks are important because a quality client that cannot balance synthetic ledgers will not support realistic reconciliation testing.

Provisioning time was measured as synthetic pipeline runtime for the simulated dataset size. The production clone took 36.0 hours because it included full refresh, copy, and post-copy manual restriction steps. Static masking took 18.5 hours. Rules-only synthesis took 12.0 hours, CTGAN synthetic generation took 14.8 hours, and the hybrid baseline took 9.4 hours. SEQ-FCT took 6.7 hours because scenario expansion was applied selectively and validation gates were automated. The runtime values are synthetic engineering estimates rather than measured enterprise deployment timings. They are included to show that secure provisioning can be evaluated alongside privacy and utility rather than treated as a separate operations concern.

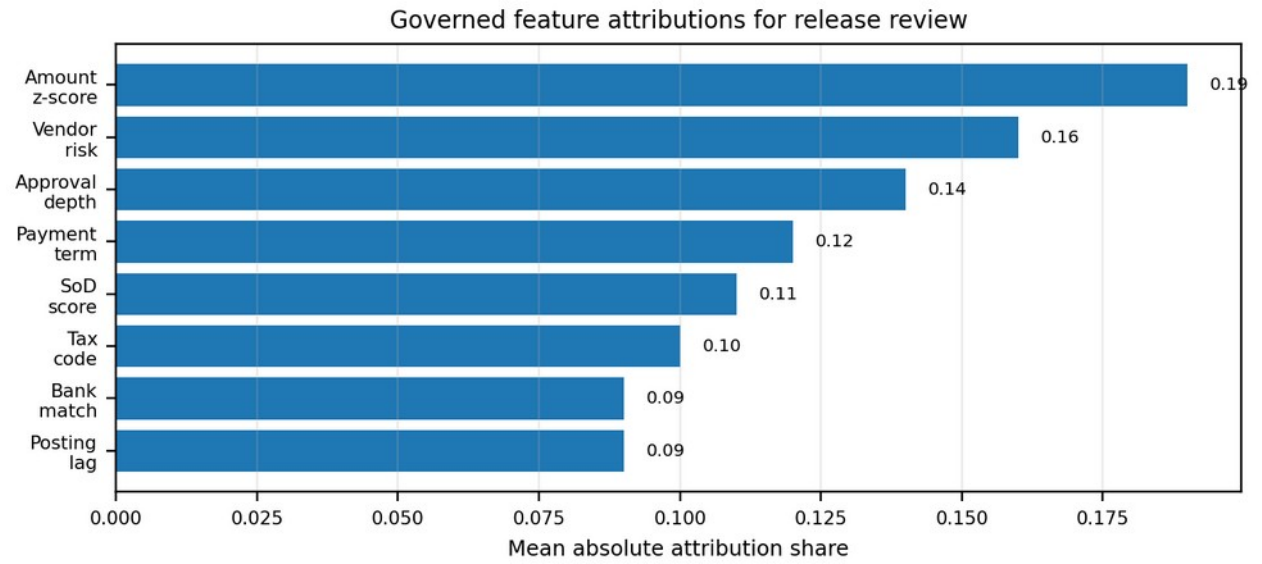

Fig. 5. Mean absolute attribution share for governed features used in release review.

*Source/derivation: Attribution shares computed from the fitted synthetic validation model.*

The explainability output in Fig. 5 shows that the highest attribution shares for control outcomes came from amount z-score, vendor-risk score, approval depth, payment term, role-conflict score, tax code, bank-match status, and posting lag. This pattern is plausible for financial control testing because it emphasizes business variables rather than masked identifiers. The governance review would flag a release if identifier tokens or free-text artifacts became dominant predictors, since that would indicate possible leakage or overfitting to transformed keys. In the synthetic experiment, attribution review supported release acceptance because the top features matched finance-control expectations and the leakage-risk score remained below the predefined threshold. Sensitivity analysis also indicated that removing targeted synthetic expansion reduced fraud recall by 4.6 percentage points, while removing relationship-aware amount transformation reduced reconciliation F1 by 3.9 percentage points. These ablation results suggest that the gain is distributed across pipeline components rather than concentrated in a single model choice.

## VI. DISCUSSION

The results suggest that secure ERP data provisioning should be evaluated as a control-testing pipeline, not merely a privacy tool. Masking alone can lower direct exposure, but it may also destroy the relationships needed to validate reconciliations, fraud rules, and audit analytics. Synthetic data alone can lower exposure further, but generated rows may miss the process variants that make controls meaningful. SEQ-FCT performed better in the synthetic evaluation because it combined deterministic relationship preservation, control-aware synthetic expansion, leakage scoring, task-based validation, and auditable release governance. The framework therefore connects privacy engineering with finance testing outcomes.

Practical implementation would require several organizational choices. Finance control owners would need to define minimum utility thresholds and critical scenarios. Privacy and security teams would need to define acceptable leakage-risk thresholds, retention periods, and access conditions. Data engineering teams would need to maintain tokenization maps, schema metadata, lineage, and transformation recipes. Internal audit would need to review release evidence and evaluate whether quality data can support reliance on automated test results. This distribution of ownership mirrors the fact that ERP data are both technical assets and financial-control evidence.

Several limitations must be stated clearly. The dataset is synthetic and intentionally structured to make all reported values internally consistent. Real ERP landscapes contain legacy custom fields, incomplete documentation, unstructured attachments, inconsistent master data, one-off process variants, and jurisdiction-specific rules that are not fully represented here. The leakage-risk score is a practical proxy, not a formal privacy proof. Differential privacy was not applied to the final dataset, and no membership-inference attack was executed against a real production distribution. The validation tasks are aligned to reconciliation, fraud-rule, and control-failure labels generated by the synthetic rules. Production deployment would require calibration against known control failures, real audit findings, and current regulatory obligations.

The framework also carries operational risks. Deterministic tokenization can become a sensitive asset if token maps are poorly protected. Scenario expansion may create unrealistic cases if finance experts do not review rule constraints. Automated approval can become a rubber stamp if thresholds are not periodically recalibrated. Data refreshes can also cause hidden drift when configuration, chart of accounts, payment terms, tax rules, or bank interfaces change. These risks do not invalidate the proposed method, but they indicate that secure provisioning must remain part of continuous governance rather than a one-time technical project. Practical deployment should therefore include separation of duties for token-map custody, restricted access to transformation configurations, periodic privacy reviews, and exception handling for cases where production examples are legally or operationally unavailable. Monitoring should also compare each refresh with previous quality releases so that sudden changes in control coverage, leakage score, or data shape are reviewed before testers rely on the environment.

## VII. CONCLUSION AND FUTURE WORKS

This work presented Secure ERP Quality Provisioning for Financial Control Testing, a governed framework for producing non-production ERP finance data that preserves control-testing utility while reducing exposure of sensitive values. The framework integrates metadata-driven sensitivity classification, deterministic tokenization, control-aware masking, targeted synthetic scenario expansion, automated validation, explainability review, and release evidence. A single synthetic dataset containing 186,000 finance-process records across six subsidiaries and five process families was used throughout the evaluation. The dataset was explicitly synthetic, and the results should be interpreted as controlled evidence rather than production validation.

The experimental analysis compared SEQ-FCT with production cloning, static masking, rules-only synthetic generation, CTGAN-style synthesis, and a hybrid baseline. The proposed framework achieved a reconciliation F1 score of 0.932, fraud-trigger recall of 0.887, control-failure F1 of 0.914, and leakage-risk score of 0.018. These results show that strong utility and low estimated leakage can coexist when transformation rules are tied to ERP relationships and financial-control tasks. The analysis also showed that utility cannot be inferred from superficial realism. Some generated datasets preserved distributions but failed to preserve edge cases needed for control testing. Conversely, some masked datasets looked safe but broke ledgers, approval patterns, or duplicate-detection logic.

The practical implication is that ERP quality-data release should be treated as a governed financial-control activity. A release package should include source scope, purpose, policy version, sensitivity classification coverage, tokenization method, synthetic-scenario inventory, referential-integrity checks, reconciliation metrics, fraud-rule metrics, leakage-risk score, attribution review, exceptions, and approver identity. Such evidence can help internal audit, privacy teams, and finance control owners evaluate whether non-production testing is both useful and appropriately protected. The framework is intended to support this evidence-driven operating model.

Future work should extend the synthetic evaluation in four directions. First, formal privacy evaluation should be added through membership-inference, attribute-inference, and linkage-risk testing, with optional differential privacy for high-risk release types. Second, process mining should be integrated more deeply so that event-order similarity, variant coverage, and control-flow deviations become release metrics. Third, real-world pilot studies should evaluate the framework in controlled enterprise settings with de-identified configurations, active audit oversight, and post-release monitoring. Fourth, governance automation should be connected to identity, access-management, and data-catalog platforms so that release approvals, retention rules, and access reviews are enforced continuously.

The conclusion is intentionally bounded. SEQ-FCT does not eliminate the need for legal review, security controls, segregation of duties, or audit judgment. It also does not prove that synthetic data are always safer or more useful than masked data. The contribution is a structured way to combine masking, synthesis, validation, and governance so that ERP quality data can support reconciliation, fraud-rule, and audit testing with lower exposure risk. Real adoption will depend on domain-specific calibration, careful token-map protection, and sustained involvement from finance, security, privacy, data engineering, and audit stakeholders. Future implementation work should define standard control libraries for common ERP finance scenarios, including three-way match, duplicate invoice review, payment-block handling, manual journal approval, bank reconciliation, tax exception handling, and intercompany clearing. Such libraries would make quality-data release more repeatable and would allow organizations to compare data refreshes across projects. A second future direction is integration with continuous controls monitoring, where validation evidence from quality environments is linked to production control owners without transferring sensitive production values. This would help close the gap between development testing and audit assurance.